\pdfoutput=1

\documentclass[preprint,12pt]{elsarticle}

\usepackage{pifont}
\usepackage{times}
\usepackage{latexsym}
\usepackage{amssymb}
\usepackage{amsmath}
\usepackage{booktabs}
\usepackage{hyperref}
\usepackage{multirow}
\usepackage{graphicx}
\usepackage{subfigure}
\usepackage{enumitem}
\usepackage{xcolor}
\usepackage{xspace}
\usepackage{comment}
\usepackage{etoolbox,caption,subcaption}
\usepackage{microtype}

\usepackage[T1]{fontenc}

\usepackage[utf8]{inputenc}

\usepackage{microtype}

\usepackage{inconsolata}

\usepackage{graphicx}

\usepackage{color}

\journal{Knowledge-Based Systems}

\begin{document}

\begin{frontmatter}


\title{Knowledge Editing for Large Language Model with Knowledge Neuronal Ensemble}


\author[NEU1]{Yongchang Li} 
\ead{lizhixi212@gmail.com}
\author[Tsinghua-ZJ]{Yujin Zhu}
\ead{zhuyujin@tsinghua-zj.edu.cn}
\author[NEU1]{Tao Yan}
\ead{2401945@stu.neu.edu.cn}
\author[Tsinghua-ZJ]{Shijian Fan}
\ead{fanshijian@tsinghua-zj.edu.cn}
\author[NEU1,NEU2]{Gang Wu\ding{41}}
\ead{wugang@mail.neu.edu.cn}
\author[Tsinghua-ZJ]{Liang Xu\ding{41}}
\ead{xlpaul@126.com}

\affiliation[NEU1]{organization={School of Computer Science and Engineering, Northeastern University},
            city={Shenyang},
            postcode={110819}, 
            country={China}}
\affiliation[NEU2]{organization={Key Laboratory of Intelligent Computing in Medical Image, Ministry of Education, Northeastern University},
            city={Shenyang},
            postcode={110819}, 
            country={China}}
\affiliation[Tsinghua-ZJ]{organization={Artificial Intelligence Innovation Center, Yangtze Delta Region Institute of Tsinghua University},
            state={Zhejiang},
            postcode={314006}, 
            country={China}}

\begin{abstract}

As real-world knowledge is constantly evolving, ensuring the timeliness and accuracy of a model's knowledge is crucial. This has made knowledge editing in large language models increasingly important. However, existing knowledge editing methods face several challenges, including parameter localization coupling, imprecise localization, and a lack of dynamic interaction across layers. In this paper, we propose a novel knowledge editing method called Knowledge Neuronal Ensemble (KNE). A knowledge neuronal ensemble represents a group of neurons encoding specific knowledge, thus mitigating the issue of frequent parameter modification caused by coupling in parameter localization. The KNE method enhances the precision and accuracy of parameter localization by computing gradient attribution scores for each parameter at each layer. During the editing process, only the gradients and losses associated with the knowledge neuronal ensemble are computed, with error backpropagation performed accordingly, ensuring dynamic interaction and collaborative updates among parameters. Experimental results on three widely used knowledge editing datasets show that the KNE method significantly improves the accuracy of knowledge editing and achieves, or even exceeds, the performance of the best baseline methods in portability and locality metrics.
\end{abstract}

\end{frontmatter}

\section{Introduction}
Real-world knowledge is constantly evolving, and the purpose of knowledge editing\cite{wang-2024-ke-survey} in large language models is to modify outdated or incorrect knowledge with new, accurate knowledge while minimizing negative effects on previously learned knowledge and capabilities. Current update methods for large language models include fine-tuning\cite{han2024parameterefficientfinetuninglargemodels} and retrieval augmentation\cite{gao2024retrievalaugmentedgenerationlargelanguage}.Fine-tuning requires high computational resources, is prone to over-fitting, may negatively impact other knowledge, and often leads to catastrophic forgetting. Retrieval augmentation, on the other hand, struggles with retrieval noise, making precise editing difficult, and provides only short-term, temporary changes, limiting its efficiency for large-scale updates. Knowledge editing\cite{yao-etal-2023-editing} seeks to empower large language models to learn continuously and maintain accurate knowledge, much like humans who read books and newspapers daily.

Existing knowledge editing methods generally involve two main steps\cite{zhang2024comprehensivestudyknowledgeediting}: parameter localization and parameter editing. Parameter localization serves as the foundation for understanding the internal working mechanisms of the model and for performing effective parameter editing. However, current knowledge editing methods face several challenges in both localization and editing: (1) \textbf{Knowledge Localization Coupling}:Knowledge localization is often coupled, meaning a single neuron may correspond to multiple pieces of knowledge, leading to frequent parameter adjustments that may destabilize the model or degrade the quality of specific knowledge edits, ultimately compromising overall performance. (2) \textbf{Inaccurate Knowledge Localization}:Current localization techniques may be inaccurate, diminishing the specificity and efficiency of the editing process. For instance, causal tracking methods may reveal significant associations between certain layers and specific knowledge, even when those layers are not directly edited. (3) \textbf{Insufficient Layer-wise Dynamic Interaction for Parameter Update}:Furthermore, when editing parameters from shallow to deep layers, dynamic interaction and collaborative updates between layers are often lacking, which can negatively impact the final model. Therefore, optimizing the knowledge editing framework is crucial for improving its effectiveness.

Inspired by knowledge neuron\cite{kn} method, we introduce a novel knowledge editing method:\textbf{K}nowledge \textbf{N}euronal \textbf{E}nsemble (\textbf{KNE}). A knowledge neuronal ensemble is a collection of neurons that represents a set of related knowledge, addressing the problem of frequent parameter modification caused by knowledge localization coupling. In this method, we calculate gradient attribution scores for each parameter in each layer to identify all parameters that significantly contribute to representing specific knowledge, allowing for more accurate and refined parameter localization. During the editing process, gradients and losses over the knowledge neuronal ensemble are computed, and error backpropagation is applied, ensuring dynamic interaction and collaborative updates among the edited parameters. Experiments conducted on three widely used knowledge editing datasets demonstrate that KNE achieves superior performance. Compared to five baseline methods, KNE significantly improves the accuracy of knowledge editing and matches or exceeds the best baseline methods in portability and locality metrics, with some datasets showing even better results.

Moreover, experiments indicate that editing the knowledge neuronal ensemble corresponding to the key layers of the feed-forward neural network (FFN) yields results comparable to those from previous methods that edited value layers, with even better locality performance. This finding enriches existing assumptions about knowledge storage 
\cite{ffn-kv1,ffn-kv2} locations and suggests that different parameter locations can be edited based on the specific editing goal to achieve optimal results.

The contributions of this paper are as follows:
\begin{itemize}
\item \textbf{Knowledge Neuronal Ensemble Method}:
The concept of the "knowledge neuronal ensemble" is introduced, solving the problem of frequent parameter modification due to knowledge localization coupling.

\item \textbf{More Precise and Accurate Knowledge Localization}: 
Gradient attribution scores are used to identify the parameters that have a significant impact on expressing specific knowledge, ensuring more accurate localization.

\item \textbf{Layer-wise Dynamic Interaction for Parameter Update}: 
Gradients and losses over the knowledge neuronal ensemble are used for error backpropagation, facilitating dynamic information transfer across layers for collaborative parameter updates.

\item \textbf{Low Computational Cost and Efficient Localization}: 
By optimizing the update strategy, the number of parameters that need to be modified is reduced to around 1\% of the model’s total parameters, significantly reducing computational cost and improving localization efficiency.
\end{itemize}

\section{Related Work}
Existing knowledge editing methods can be broadly categorized into two types\cite{yao-etal-2023-editing}: those that modify model parameters and those that do not. 

Among the non-parameter-modifying methods, there are two paradigm. The first is knowledge editing based on retrieval augmentation, which treats the new knowledge as external knowledge in the retrieval-augmented model,i.e.SERAC\cite{serac}, IKE\cite{ike}, Wang et al.\cite{wang2023retrievalaugmented}, Shi et al.\cite{shi2024retrievalenhancedknowledgeeditinglanguage}, MemPrompt\cite{MemPrompt}, Murty et al.\cite{murty2022patches}. 
The second involves adding extra trainable parameters, which are trained on the modified knowledge dataset while the original model parameters remain unchanged,i.e.T-Patcher\cite{Transformer-Patcher}, CaliNet\cite{Calibrat}, GRACE\cite{GRACE}, MELO\cite{melo}. Both of them leave the original model parameters unchanged, leading to the model’s inability to deeply understand or fully integrate the new knowledge. 

Although parameter-modifying methods are more challenging, they facilitate a deeper understanding of knowledge storage within the model's internal mechanisms. This enables the model to better grasp the inherent nature of knowledge and apply it flexibly, a focus of our research. Parameter-modifying methods can also be classified into two types: meta-learning-based methods and locate-then-edit methods.

Meta-learning methods use hyper networks for learning parameter updates for large language models (LLMs). The Knowledge Editor (KE)\cite{decao2021editing}utilizes a hyper network (specifically a bidirectional LSTM) to predict the weight updates for each data point, thereby achieving constrained optimization in editing target knowledge without disrupting other knowledge. However, this approach faces limitations when editing LLMs. To overcome this, Model Editing Networks with Gradient Decomposition (MEND)\cite{mend}learns to transform the fine-tuning gradients of language models through low-rank decomposition, achieving better performance on LLMs. MALMEN\cite{MALMEN} formulates parameter shift aggregation as a least squares problem and updates model parameters using the normal equation, allowing for scalable editing of multiple facts with limited memory.

The locate-then-edit approach first identifies the parameters corresponding to specific knowledge and modifies them by directly updating the target parameters. The Knowledge Neuron (KN) method\cite{kn} introduces a knowledge attribution technique to locate "knowledge neurons" (key-value pairs in the FFN matrix) that embody the knowledge, and subsequently updates these neurons. ROME\cite{rome}employs causal mediation analysis to locate the editing region. Unlike modifying knowledge neurons in the FFN, ROME adjusts the entire matrix, viewing model editing as a least-squares problem with linear equality constraints, solving it using Lagrange multipliers. However, both KN and ROME can only edit one factual association at a time. To address this, MEMIT\cite{memit}extends ROME, enabling simultaneous editing of multiple cases. Building on MEMIT, PMET\cite{li2023pmet}introduces attention values for better performance.

\section{Method}
\subsection{Preliminaries}
As in Wang et al.\cite{wang-2024-ke-survey}, the editing is performed on a relational fact that can be represented as a knowledge triple $(s, r, o)$ where $s$, $r$, and $o$ are the subject, relation, and object of the fact respectively. A single knowledge editing task denoted by $e$ is to modify the weights of the model such that the original knowledge triple encoded in the model is changed to $(s, r, o^*)$. Since the context of this work is pre-trained LLMs, we use $x_e = (s, r)$ to represent the prompt composed of $s$ and $r$, and substitute $y_e$ for $o$ to represent the answer. Thus, a LLM with parameter $\phi$ can be regarded as a mapping represented by $f: x \rightarrow y$, and a knowledge editing task $e$ will yield $f^*: x \rightarrow y^*$ with the parameter updated to $\phi^*$. It is more common to perform multiple knowledge editing on a set of edits $\mathcal{E} = \{e_1, e_2, \ldots\}$. Let $\mathcal{X_E} = \cup_{e \in \mathcal{E}} x_e$ and $\mathcal{Y_E} = \cup_{e \in \mathcal{E}} y_e$. We formalize the problem of \textbf{Knowledge Editing (KE)} following the definition in Wang et al.\cite{wang-2024-ke-survey}.

\textbf{Definition 1} The objective of knowledge editing is to perform the following constrained optimization:
\begin{equation}
\begin{aligned}
&\min_{\phi^*} \mathbb{E}_{e \in \mathcal{E}} \mathbb{E}_{x\in \mathcal{X}_e, y^* \in \mathcal{Y}^*_e} \mathcal{L}(f^*(x), y^*),
\\
&\mathrm{s.t.}\ f^*(x) = f(x), \quad \forall x \in \mathcal{X} \setminus \mathcal{X}_E,
\end{aligned}
\end{equation}
where $\mathcal{L}$ is the loss function which quantifies the deviation between the LLM output $f^*(x)$ and $y^*$ from the expected answer set $\mathcal{Y}^*_e$.

For methods of Locate-Then-Edit paradigm, locating neurons attributed to the specific knowledge triple, i.e., Knowledge Attribution, is a prerequisite. We refine the problem statement of knowledge editing in this case.

\textbf{Definition 2} The objective of Locate-Then-Edit knowledge editing is to perform the following constrained optimization:
\begin{equation}
\begin{aligned}
&\min_{\phi^*_k} \mathbb{E}_{e \in \mathcal{E}} \mathbb{E}_{x\in \mathcal{X}_e, y^* \in \mathcal{Y}^*_e} \mathcal{L}\left(f^*_{\bar{\phi}_k, \phi^*_k}(x), y^*\right),\\
&\mathrm{s.t.}\ f^*_{\bar{\phi}_k, \phi^*_k}(x) = f(x), \quad \forall x \in \mathcal{X} \setminus \mathcal{X}_E,\\
&\mathrm{where}\ \phi_k = L(f_\phi, \mathcal{E}), \quad \bar{\phi}_k = \phi \setminus \phi_k.\\
\end{aligned}
\end{equation}
Here, $\phi_k = L(f_\phi, \mathcal{E})$ uses function $L$ to locate neurons attributed to edits $\mathcal{E}$. Then, $\bar{\phi}_k$ represents the unedited weights.

\textbf{Definition 3 (Knowledge Neurons, KNs)}. Given a set of edits $\mathcal{E}=\{e_1, \ldots\}$ where $|\mathcal{E}|\geq 1$, for a $L$ layer neural network, suppose there are $m$ neurons in layer $l$. If $\mathcal{E}$ can be attributed to $k$ neurons $w_{j_1}^{(l)}, \ldots, w_{j_k}^{(l)}$ in layer $l$, such that the activation of these neurons significantly contributes to $\mathcal{E}$, then these neurons $\mathbf{N}_{\mathcal{E}}^{(l)}=\{w_{j_1}^{(l)}, \ldots, w_{j_k}^{(l)}\}$ are referred to as the knowledge neurons for $\mathcal{E}$ in layer $l$.

\textbf{Definition 4 (Knowledge Neuronal Ensemble, KNE)}. The Knowledge Neuronal Ensemble of the $L$ layers neural network for $\mathcal{E}$ is defined to be the set of knowledge neurons in all layers, i.e., $\mathbf{N}_\mathcal{E}=\{\mathbf{N}_\mathcal{E}^{(1)}, \ldots, \mathbf{N}_\mathcal{E}^{(l)}\}$.
\begin{figure*}[!ht]
    \centering
    \includegraphics[width=\textwidth]{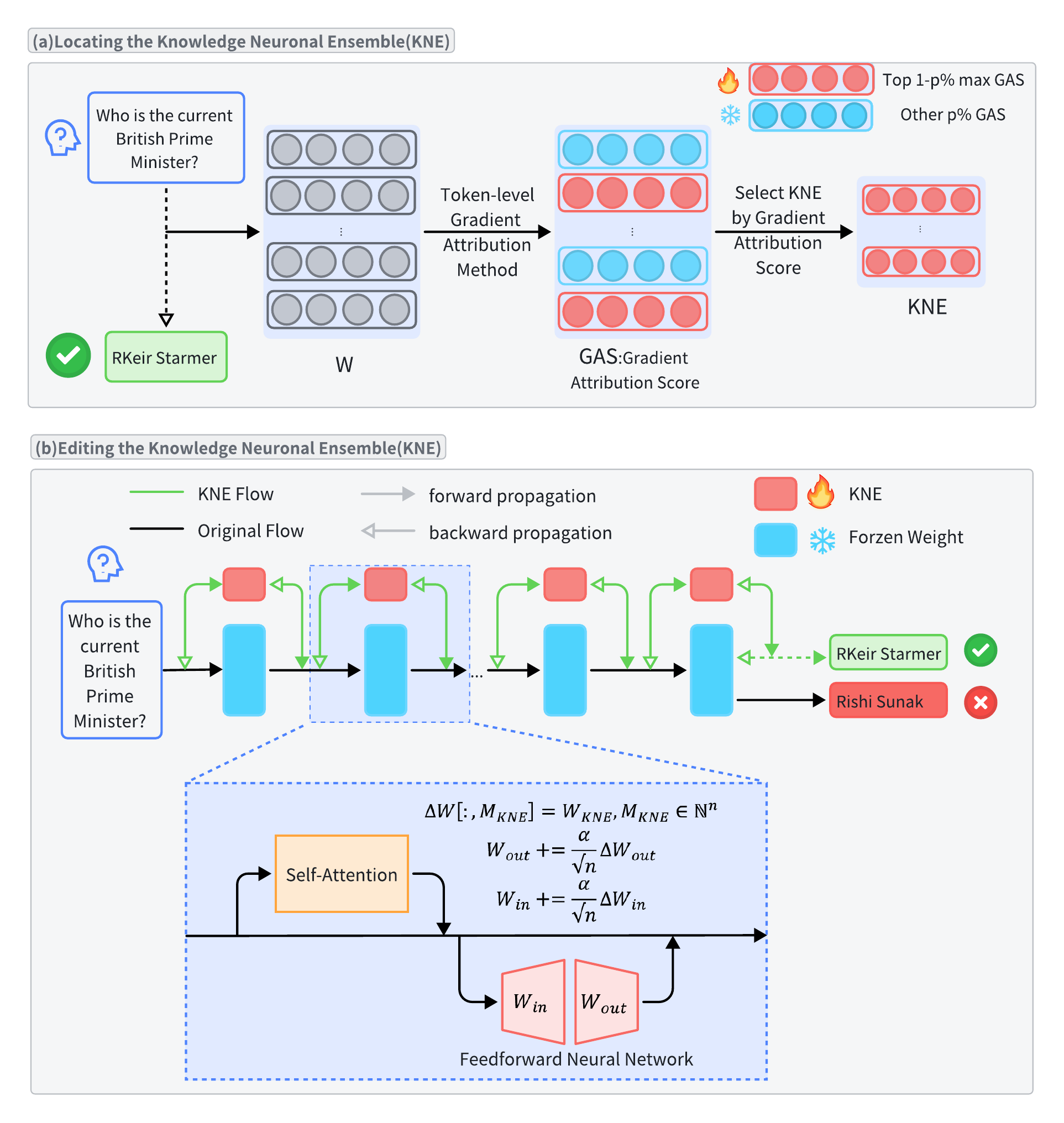}
    \caption{The framework of KNE method.}
    \label{fig:fullpageimage}
\end{figure*}
\subsection{Knowledge Neuronal Ensemble Method}
To localize the Knowledge Neuronal Ensemble corresponding to a set of knowledge, we use a token-level gradient attribution method\cite{kn} to calculate gradient attribution scores for the relevant parameters. Based on these scores, we select and construct the Knowledge Neuronal Ensemble. Afterward, we freeze the parameters in other locations and dynamically update only the parameters in the Knowledge Neuronal Ensemble to achieve coordinated optimization.The framework of the Knowledge Neuronal Ensemble is shown in Figure\ref{fig:fullpageimage}.

\subsubsection{Token-level Gradient Attribution Method}
For a given input query \( x \), and the correct answer \( \mathrm{y_{j}^{*}} = \{y_1, y_2, \ldots, y_j\} \), where \( \mathrm{y_{j}^{*}} \) represents the correct answer composed of \( j \) tokens and \( y_j \) represents the \( j \)-th token, we define the model's output probability as:
\begin{equation}
\mathrm{P}_{\mathrm{y_{j}^{*}}|x}(\hat{w}_{k}^{(l)}) = p(\mathrm{y_{j}^{*}}|x, w_{k}^{(l)} = \hat{w}_{k}^{(l)})
\end{equation}
Here, \( x \) is the input query, \( \mathrm{y_{j}^{*}} \) is the correct answer, \( w_{k}^{(l)} \) refers to the \( k \)-th neuron in the \( l \)-th layer, and \( \hat{w}_{k}^{(l)} \) represents the parameter value at location \( w_{k}^{(l)} \).

To extend the gradient attribution method to large language models built on GPT-like architectures, we accumulate the gradient attribution scores computed for each token in the correct answer:
\begin{equation}
\mathrm{Attr}(w_{k}^{(l)}) = \overline{w}_{k}^{(l)} \sum_{j=1}^{s} \int_{\alpha=0}^{1} \frac{\partial \mathrm{P}_{\mathrm{y_{j}^{*}}|x}(\alpha \overline{w}_{k}^{(l)})}{\partial w_{k}^{(l)}} d\alpha
\end{equation}
Where \( s \) is the number of tokens in the correct answer, and \( \overline{w}_{k}^{(l)} \) is the original parameter value at location \( w_{k}^{(l)} \).

Since calculating the continuous gradient integral is computationally expensive, we apply a Riemann approximation as follows:
\begin{equation}
\widetilde{\mathrm{Attr}}(w_{k}^{(l)}) = \frac{\overline{w}_{k}^{(l)}}{m} \sum_{j=1}^{s} \sum_{i=1}^{m} \frac{\partial \mathrm{P}_{\mathrm{y_{j}^{*}}|x}\left(\frac{i}{m} \overline{w}_{k}^{(l)}\right)}{\partial w_{k}^{(l)}}
\end{equation}
Where \( m \) represents the number of discrete steps used to approximate the continuous integral, reducing computational complexity.

\subsubsection{Selection of the Knowledge Neuronal Ensemble}
We define \( \mathrm{Attr}(w_{k}^{(l)}) \) as the gradient attribution score of the \( k \)-th neuron in the \( l \)-th layer, and let \( d_2 \) represent the output dimension of the \( l \)-th layer. To construct the Knowledge Neuronal Ensemble (KNE), we select the neurons with the top \(1-p\%\) largest gradient attribution scores. The threshold \( t_p \) is defined as follows:
\begin{equation}
\begin{split}
t_p &= \mathrm{Quantile}_{1-p}(\{\mathrm{Attr}(w_{k}^{(l)}) \mid \\
    & \quad k = 1, 2, \dots, d_2 ;\ l=1, 2, \dots, L\}),
\end{split}
\end{equation}
where \( \mathrm{Quantile}_{1-p} \) refers to the quantile function that calculates the value corresponding to the top \( p\% \) of the gradient attribution scores.

Next, we define the Knowledge Neuronal Ensemble (KNE) as the set of neuron indices that meet the following condition:
\begin{equation}
\begin{split}
\mathbf{N}_\mathcal{E}^{*} &= \{\{k^{(l)}\} \mid \mathrm{Attr}(w_{k}^{(l)}) \geq t_p, \\
    & \quad k = 1, 2, \dots, d_2; l = 1, 2, \dots, L\}.
\end{split}
\end{equation}
In this way, the KNE includes all neuron indices with gradient attribution scores greater than or equal to the threshold \( t_p \), representing the top \(1-p\%\) of neurons ranked by gradient attribution scores.

\subsubsection{Editing the Knowledge Neuronal Ensemble}
To fully utilize the localized information from the Knowledge Neuronal Ensemble (KNE), we propose a Knowledge Neuronal Ensemble Editing method. Throughout the editing phase, gradients and losses are calculated across the knowledge neuronal ensemble. Subsequently, error backpropagation is employed to facilitate dynamic interaction and coordinated updates among the parameters being refined. Based on the number of neurons \( n \) in the Knowledge Neuronal Ensemble for each layer, we dynamically allocate a Knowledge Neuronal Ensemble parameter matrix: 
\begin{equation}
W_{kne} \in \mathbb{R}^{n \times d_1} 
\end{equation}
Here, \( n \) represents the number of neurons in the Knowledge Neuronal Ensemble for that layer, while \( d_1 \) and \( d_2 \) represent the input and output dimensions of the weight matrix \( W \). 
Since \( W_{kne} \) and \( W \) have different dimensions, we initialize a zero matrix \( \Delta W \) with the same dimensions as \( W \), to map the updated \( W_{kne} \) to the corresponding positions of the Knowledge Neuronal Ensemble. The formulation is as follows: 
\begin{equation}
\Delta W \in \mathbb{R}^{d_2 \times d_1}, W \in \mathbb{R}^{d_2 \times d_1} 
\end{equation}
\begin{equation}
\Delta W[:, M_{kne}] = W_{kne}, M_{kne} \in \mathbb{N}^{n} 
\end{equation}
Where \( M_{kne} \) represents the index positions of the Knowledge Neuronal Ensemble within the weight matrix \( W \). The values of \( M_{kne} \) are natural numbers less than \( d_2 \), and the length of \( M_{kne} \) corresponds to the number of neurons \( n \) in the Knowledge Neuronal Ensemble. 
Finally, the weight matrix \( W \) is updated to obtain \( \hat{W} \) using the following formula: 
\begin{equation}
\hat{W} = W + \frac{\alpha}{\sqrt{n}} \Delta W 
\end{equation}
To control the extent of the parameter updates, we multiply \( \Delta W \) by a scaling factor \( \frac{\alpha}{\sqrt{n}} \), where \( \alpha \) is a hyper parameter.


\section{Experiments}
\subsection{Experimental Setting}
In our study, we chose the \textbf{Llama2-7B-chat} and \textbf{gpt-j-6B} models as the cornerstone for knowledge editing tasks. These models were selected for their widespread adoption and proven efficacy. To ensure a comprehensive evaluation, we employed diverse datasets that encapsulate a broad spectrum of knowledge domains, namely the \textbf{ZsRE}, \textbf{WikiDatacounterfact}, and \textbf{WikiDatarecent} datasets. Our assessment criteria encompassed several key metrics: \textbf{Edit Success}, which measures the accuracy of edits; \textbf{Portability}, indicating how well edits transfer across different questions; \textbf{Locality}, assessing the precision of edit localization; and \textbf{Fluency}, evaluating the naturalness of the edited text. We benchmarked our approach against established baseline methods, comprising Fine-Tuning (\textbf{FT}), Fine-Tuning with Linear probing (\textbf{FT-L}), \textbf{AdaLoRA}, \textbf{ROME}, and \textbf{MEMIT}. For an in-depth exploration of these comparisons and additional methodological details, please refer to \ref{sec:appendix}.

\subsection{Experimental Results}


We selected the widely used \textbf{Llama2-7B-chat} model as the foundation for knowledge editing. To validate the generalizability of our approach, we utilized datasets that represent various forms of knowledge, including \textbf{ZsRE} and \textbf{WikiDatacounterfact}. Additionally, to assess performance over time, we incorporated the \textbf{WikiDatarecent} dataset. For comparison, we included several previously successful knowledge editing methods, with the experimental results summarized in Table \ref{Tab:PerformanceComparison}.

Furthermore, to evaluate the effectiveness of our method across different models, we compared it with the \textbf{gpt-j-6B} model. The \textbf{Llama2-7B-chat} model demonstrated exceptional performance in handling complex knowledge editing tasks due to its larger parameter size and enhanced generative capabilities. In contrast, the \textbf{gpt-j-6B} model is favored for its lower computational resource requirements and faster response times. By comparing these two models, we gained deeper insights into their respective strengths and limitations in knowledge editing tasks. The results of these comparative experiments are summarized in Table \ref{Tab:PerformanceComparison2}.

\begin{table*}[!h]
	\centering
	\caption{Performance Comparison of Knowledge Editing Methods Across Different Datasets}
	\resizebox{\linewidth}{!}{
		\begin{tabular}{ccccccccc}
			\toprule
			Dataset & {Metric}  & \textbf{FT}   & \textbf{FT-L} & \textbf{AdaLoRA} & \textbf{ROME} & \textbf{MEMIT} & KNE \\ \midrule
			\multirow{4}{*}{WikiData counterfact} & Edit Succ. & 26.78 & 51.12 & 72.14 & 83.21 & 83.41 & \textbf{99.02} \\ 
			& Portability & 16.94 & 39.07 & \textbf{55.17} & 38.69 & 40.09 & 53.88 \\ 
			& Locality    & 0.29  & 62.51 & \textbf{66.78} & 65.4  & 63.68 & 65.09 \\ 
			& Fluency     & 483.71 & 544.80 & 553.85 & 578.84 & 568.58 & \textbf{591.25} \\ \addlinespace
			\multirow{4}{*}{ZsRE}       & Edit Succ. & 36.88 & 54.65 & 69.86 & 96.57 & 83.07 & \textbf{97.75} \\ 
			& Portability & 8.72  & 45.02 & 52.95 & 52.20 & 51.43 & \textbf{58.02} \\ 
			& Locality    & 0.31  & 71.12 & 72.21 & 27.14 & 25.46 & \textbf{76.85} \\ 
			& Fluency     & 471.29 & 474.18 & 532.82 & 570.47 & 559.72 & \textbf{571.93} \\ \addlinespace
			\multirow{4}{*}{WikiData recent} & Edit Succ. & 31.24 & 71.18 & 65.61 & 85.08 & 85.32 & \textbf{99.48} \\ 
			& Portability & 15.91 & 48.71 & 47.22 & 37.45 & 37.94 & \textbf{63.36} \\ 
			& Locality    & 3.65  & 63.7  & 55.78 & \textbf{66.2}  & 64.78 & 37.58 \\  
			& Fluency     & 428.67 & 549.35 & 537.51 & 574.28 & 566.66 & \textbf{581.49} \\ \bottomrule
	\end{tabular}}
    \label{Tab:PerformanceComparison}

\end{table*}

\begin{table*}[!h]
    \centering
    \caption{Performance Comparison of Knowledge Editing Methods Across Different Models}
    \resizebox{\linewidth}{!}{
        \begin{tabular}{c c c c c c}
            \toprule
            Dataset                    & Model              & \textbf{Edit Succ.} & \textbf{Portability} & \textbf{Locality} & \textbf{Fluency} \\  
            \midrule
            
            \multirow{2}{*}{WikiData counterfact} 
                & Llama2-7b-chat      & 99.02      & 53.88       & 65.09   & 591.25 \\ 
                & gpt-j-6b           & 99.35      & 49.14       & 52.64   & 597.29 \\ 
            \addlinespace
            
            \multirow{2}{*}{ZsRE}  
                & Llama2-7b-chat     & 97.75      & 58.02       & 76.85   & 571.93 \\ 
                & gpt-j-6b           & 99.90      & 53.79       & 78.60   & 549.87  \\ 
            \addlinespace
            
            \multirow{2}{*}{WikiData recent} 
                & Llama2-7b-chat     & 99.48      & 63.36       & 37.58   & 581.49 \\ 
                & gpt-j-6b           & 99.79      & 57.74       & 53.47   & 585.79 \\ 
            \bottomrule
        \end{tabular}
    }
    \label{Tab:PerformanceComparison2}
\end{table*}




We evaluate several knowledge editing methods on a range of language models and datasets, assessing their performance across key metrics.

\subsection{Exploring the Storage Location of Knowledge in Large Language Models}
Exploring the storage location of knowledge in large language models has traditionally relied on the assumption that factual knowledge is stored in the \textbf{key-value memory} format within the fully connected layers of the \textbf{FFN} module, as seen in methods such as ROME, MEMIT, and the knowledge neuron approach. 
However, we have identified several interesting deviations from this assumption.

To analyze the precise locations of knowledge within these models, we employed a gradient attribution method to calculate gradient attribution scores for each parameter layer in both the Self-Attention and FFN modules. 
Using these scores, we identified specific layers and evaluated their post-editing performance, as illustrated in Figure. \ref{fig:experiments-location}. 
This analysis yielded several notable conclusions:

\begin{itemize}
    \item \textbf{Edit Success and Portability:} Editing within the FFN module consistently led to superior Edit Success and Portability compared to the Self-Attention module. 
    Notably, the value layer (\texttt{mlp.down\_proj}) in the FFN module exhibited the best overall performance. 
    However, high editing accuracy was observed across various layers and modules, indicating that effective edits are achieved throughout the model.

    \item \textbf{Locality and Fluency:} For Locality and Fluency, editing mapping layers—such as \texttt{mlp.gate\_proj} and \texttt{mlp.up\_proj} in the FFN module, along with \texttt{self\_attn.q\_proj} in the Self-Attention module—demonstrated significantly better performance than other layers.
\end{itemize}

These findings suggest that while knowledge is indeed stored in the FFN module, the specific layers edited impact different performance aspects. 
Moreover, mapping layers play a crucial role in maintaining Locality and Fluency, indicating that the storage and structure of knowledge are more complex and distributed than initially assumed.

\subsection{Similar Knowledge May Be Stored in Similar Locations within the Model}

Effective model editing does not require localizing every knowledge element in the dataset. 
By localizing only \textbf{200 knowledge items}—around \textbf{1/4} of the total dataset—the model achieved high performance. 
Remarkably, this partial localization strategy outperformed full-dataset localization in both \textbf{Locality} and overall performance. Furthermore, metrics such as \textbf{Edit Success}, \textbf{Fluency}, and \textbf{Portability} showed minimal differences between using the entire dataset and using just 1/4 for localization.

This finding significantly improves the feasibility of knowledge editing for practical applications by reducing the computational demands of the localization process. 
In the \textbf{KNE} approach, calculating gradient attribution scores for each parameter across all layers for every knowledge item is highly resource-intensive, often making localization more time-consuming than the editing itself. 
By streamlining this process and achieving a 75\% reduction in localization effort, the method accelerates overall workflow and enhances scalability for industrial applications.

The results reveal two distinct peaks in performance: 
one between \textbf{200-300} localized knowledge points and another between \textbf{700-800}. 
Localizing \textbf{200-300} knowledge points produces superior outcomes compared to \textbf{700-800}, especially regarding \textbf{Edit Success}, \textbf{Portability}, and \textbf{Locality} metrics. 
An excessive number of localized points introduces conflicts among knowledge points, diminishing overall performance.

The dataset, \textbf{WikiDatacounterfact}, derived from WikiData, contains numerous data points with inherent similarities, such as comparable classification topics Gueta et al.\cite{Gueta2023KnowledgeIA}. 
These similarities imply that related knowledge items are often stored in close model regions. 
Thus, localizing a representative subset effectively supports the editing process. 
Further experimentation is required to elucidate the underlying reasons for this behavior.

The experiment result of full dataset is shown in Figure.~\ref{fig:full_dataset} in Appendix.

\subsection{Optimal Parameter Selection for Knowledge Editing}

To investigate the impact of parameter quantity on knowledge editing performance, a controlled experiment was conducted. 
All hyperparameters were held constant except for the hyperparameter dictating the number of parameters engaged in the editing process. 
This approach isolates the effect of parameter quantity on editing efficiency and effectiveness.

\begin{itemize}
    \item \textbf{More parameters:} When a larger number of parameters are used, the \textbf{Edit Success} and \textbf{Portability} metrics show significant improvement. 
    This suggests that using more parameters allows for more precise and effective modifications to the model's knowledge, leading to better overall performance in terms of successful edits and the transfer ability of those edits across different tasks.
    \item \textbf{Fewer parameters:} When fewer parameters are used, the \textbf{Locality} metric performs better. 
    This indicates that using fewer parameters helps maintain the relevance of the edited knowledge to its surrounding context, resulting in more focused and localized edits.
\end{itemize}

The results suggest a trade-off between these two sets of metrics. 
To achieve optimal overall performance, we must balance the number of parameters used for editing. Using too many parameters enhance \textbf{Edit Succes}s and \textbf{Portability} but may negatively affect \textbf{Locality}. 
Conversely, using fewer parameters improve \textbf{Locality} but may lead to less effective edits overall. Thus, selecting an appropriate number of parameters is crucial to achieving the best overall editing performance.
Detail experiment result is shown in Figure.~\ref{fig:parameter_setting} in Appendix.

\subsection{Exploring the Capability of Batch Editing}

Most current knowledge editing methods only handle one or a few pieces of knowledge at a time, limiting the efficiency and applicability of knowledge editing. 
To evaluate whether our proposed method extend to\textbf{ batch editing}, we conducted a controlled experiment, varying only the number of knowledge pieces edited (i.e., the \textbf{batch size}). The experimental results are as follows:

\begin{figure*}[!ht]
	\centering
	\subfigure[]{\includegraphics[width=.45\linewidth]{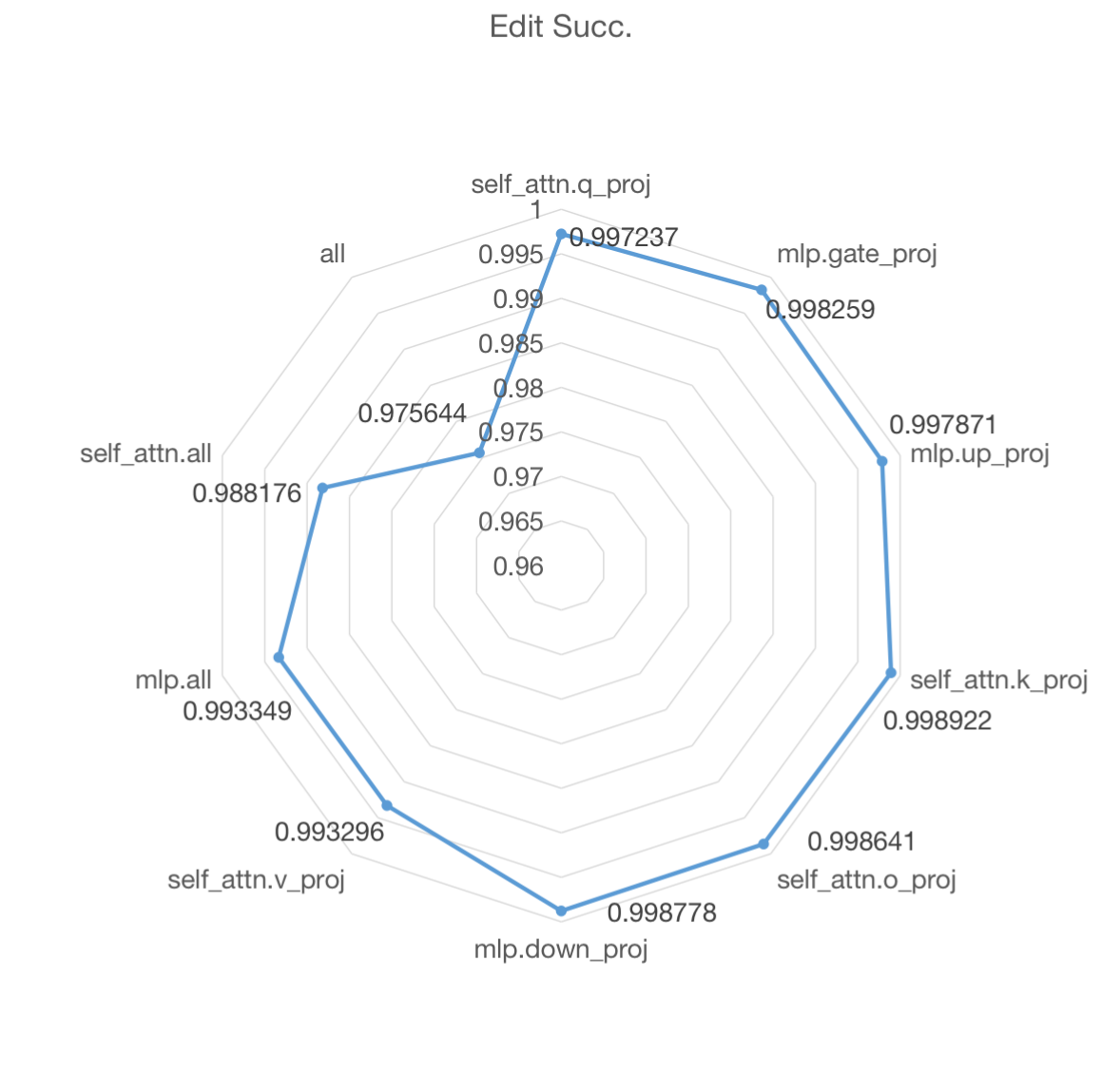}}
	\subfigure[]{\includegraphics[width=.45\linewidth]{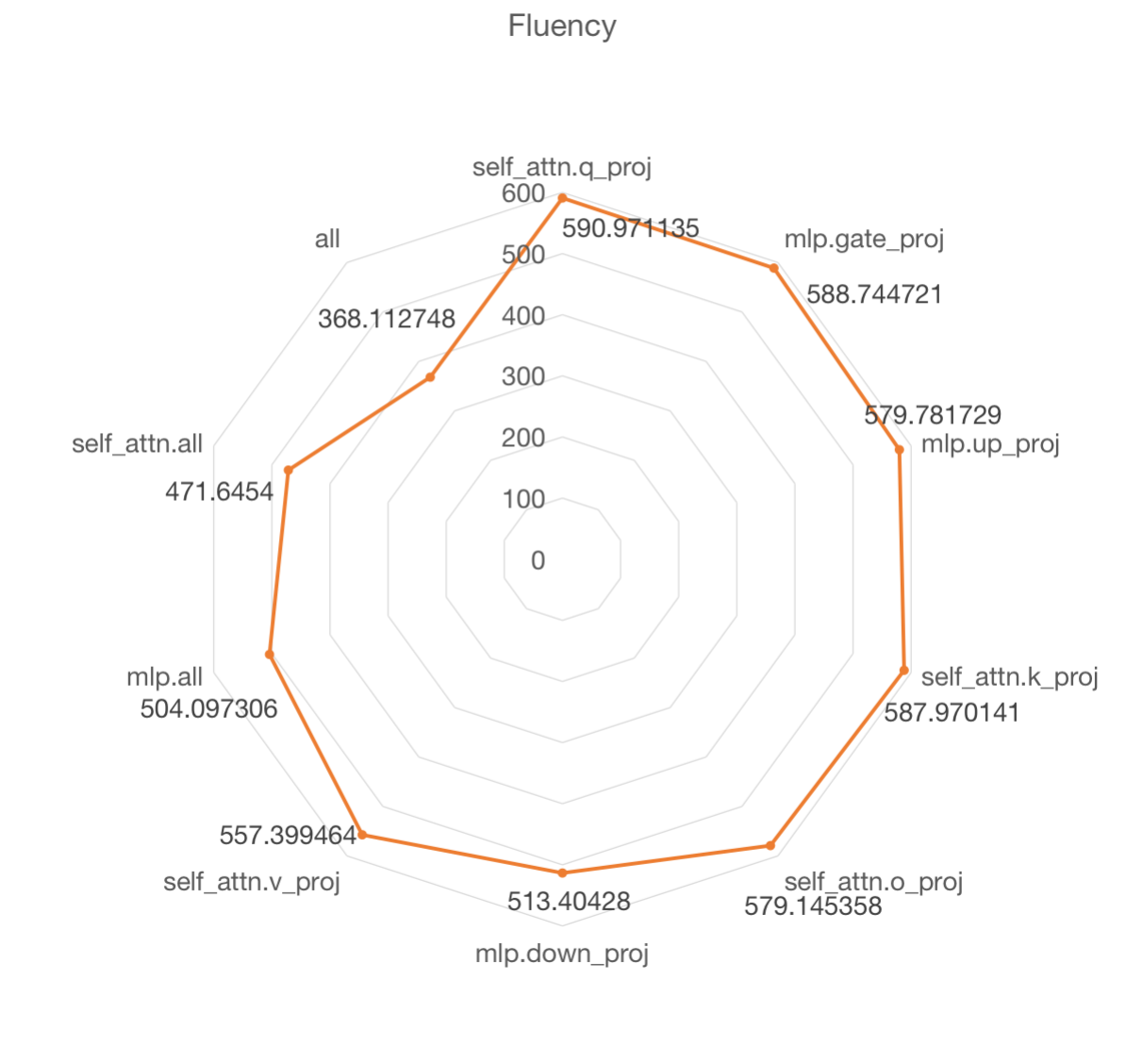}}
	\subfigure[]{\includegraphics[width=.45\linewidth]{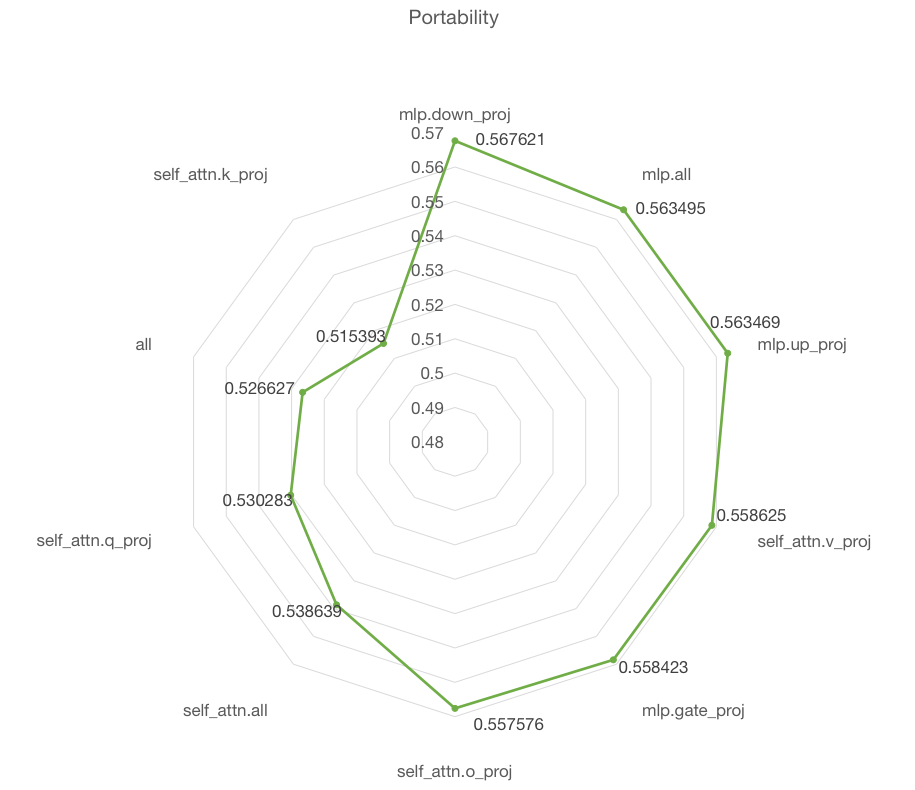}}
	\subfigure[]{\includegraphics[width=.45\linewidth]{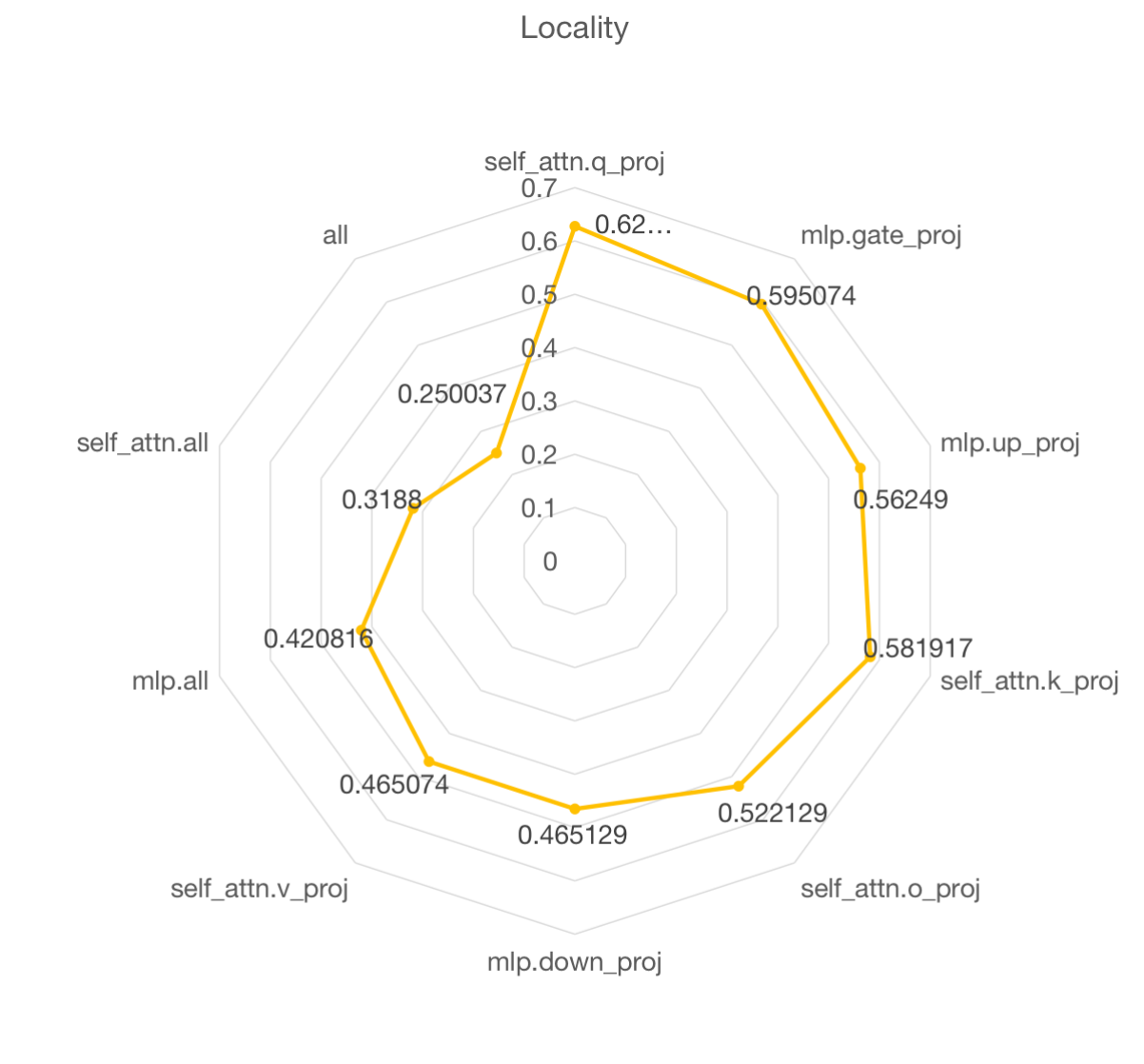}}
  \caption{Visualization of Knowledge Storage Deviations in Large Language Models}
  \label{fig:experiments-location}
\end{figure*}

The performance of our method under different batch sizes is as shown in Figure.~\ref{fig:batchsize} in Appendix.

\textbf{Edit Success and Portability}: As the \textbf{batch size} increases, the \textbf{Edit Success} and \textbf{Portability} metrics initially show a steep decline, which gradually levels off. 
Despite this reduction, the overall performance remains within an acceptable range, indicating that the method maintains Edit Success and Portability even with larger batch sizes.
\textbf{Locality}: The \textbf{Locality} metric rises sharply as the batch size increases, followed by a gradual decline. 
This suggests that batch editing initially enhances the relevance of the edited knowledge to its context, but as the batch size grows, the potential for conflicts between edits may cause a slight decrease in performance.
\textbf{Fluency}: The \textbf{Fluency} metric also increases steeply with larger batch sizes and then continues to rise more gradually. 
This indicates that batch editing improves the coherence and smoothness of the edited knowledge, though the benefits diminish as the batch size continues to grow.

Our method demonstrates the ability to perform batch knowledge editing effectively, with only minor trade-offs as batch size increases. 
While \textbf{Edit Success} and \textbf{Portability} metrics experience some decline with larger batch sizes, the performance remains acceptable. 
In contrast, \textbf{Locality} and \textbf{Fluency} improve with larger batch sizes, at least initially, showing that our method is well-suited for batch editing tasks.

\section{Conclusion}
This paper introduces a novel knowledge editing framework—the \textbf{Knowledge Neurona'l Ensemble (KNE) localization method}—to address the limitations of current knowledge editing techniques, including localization accuracy, editing efficiency, and inter-layer coordinated updates. By introducing the concept of the \textbf{Knowledge Neuronal Ensemble}, we not only expand our understanding of knowledge storage locations but also achieve more precise knowledge localization and batch updates by aggregating multiple related knowledge neurons. This method enhances inter-layer interaction through a dynamic gradient propagation mechanism from shallow to deep layers, improving the coherence and accuracy of edits while minimizing negative impacts on overall model performance.Experimental results demonstrate that the KNE localization method outperforms mainstream knowledge editing techniques across multiple datasets, delivering higher accuracy and stability in knowledge editing. It also significantly reduces computational overhead and improves localization efficiency.

\section{Discussion on Limitation}

Although the proposed \textbf{Knowledge Neuronal Ensemble (KNE) localization method} significantly improves knowledge editing accuracy, efficiency, and reduces computational overhead, several limitations remain that warrant further exploration and optimization.
\begin{itemize}
    \item \textbf{Dependency on Knowledge Set Quality}: The performance of the KNE method heavily depends on the quality of the selected knowledge set. In practical applications, ensuring that the chosen Knowledge Neuronal Ensemble is representative and avoiding over-editing or missing critical knowledge points remains a challenging task. Future research should consider incorporating more refined knowledge selection mechanisms to enhance the precision of knowledge editing while minimizing disruption to the model's existing capabilities.
    
    \item \textbf{Theoretical Explanation of Key Layer Editing}: While this paper shows that modifying the \textbf{key layer} in the FFN module yields favorable results in terms of locality, this finding still requires deeper theoretical analysis and explanation. Future studies should further investigate the knowledge transmission mechanisms between different layers of models to systematically understand and optimize knowledge storage and editing processes, thereby improving the theoretical robustness of the method.
\end{itemize}

\bibliographystyle{elsarticle-num}
\bibliography{acl_latex}

\appendix

\section{Settings}
\label{sec:appendix}

\subsection{Baselines}

We compare against the following baseline methods:

\begin{itemize}
    \item \textbf{Fine-Tuning (FT)}: We fine-tune the 'mlp.proj' weights within layer 21, following the re-implementation by Meng et al.\cite{memit}. We use the Adam optimizer with early stopping to minimize negative log probability. Default hyper parameters are used, and unconstrained fine-tuning is consistently applied across all experiments.
    \item \textbf{Fine-Tuning with Linear probing (FT-L)}: This method fine-tunes a pre-trained model by adding and training a linear layer on top, while keeping the original model weights frozen. This linear layer adapts the model to new tasks without altering the pre-trained representations. We utilize the Adam optimizer with early stopping and default hyper parameters for training.
    \item \textbf{AdaLoRA}\cite{adalora}: AdaLoRA efficiently tunes large pre-trained models by applying small, rank-1 updates to the model weights. This is particularly beneficial for large models as it significantly reduces the number of updates compared to full fine-tuning. We employ AdaLoRA on specific layers, using a variant of the Adam optimizer and default hyper parameters.
    \item \textbf{ROME}\cite{rome}: This method treats the MLP module as a key-value store and adds new knowledge via rank-one modification of MLP weights. We utilize the original code and weights (\url{https://github.com/EleutherAI/ROME}) and retain default hyper parameters.
    \item \textbf{MEMIT}\cite{memit}: An extension of ROME, MEMIT incorporates multiple memories by modifying MLP weights across several layers. We use the publicly available code (\url{https://github.com/facebookresearch/memit}) with default hyper parameters. For GPT-J, R values range from 3 to 8, and covariance statistics are derived from 100,000 Wikitext samples. 
\end{itemize}

\subsection{Models}

We evaluate the following language models:

\begin{itemize}
    \item \textbf{Llama-2-7b-chat}\cite{touvron2023llama2openfoundation, llamacode}: Meta's 7-billion parameter chat-tuned model, exhibiting strong performance in dialogue benchmarks. It utilizes an optimized transformer architecture trained with supervised fine-tuning (SFT) and reinforcement learning with human feedback (RLHF).
    \item \textbf{GPT-J-6B}\cite{gpt-j,mesh-transformer-jax}: A 6-billion parameter model with 28 layers, a model dimension of 4096, a feedforward dimension of 16384, and 16 heads (dimension 256). It uses RoPE on 64 dimensions per head and a 50257-token vocabulary with BPE encoding.
\end{itemize}

\subsection{Datasets}

We utilize the following datasets\cite{wang2023easyedit}(\url{https://huggingface.co/datasets/zjunlp/KnowEdit}):

\begin{itemize}
    \item \textbf{ZsRE}: A Question Answering (QA) dataset using back-translation paraphrases to create question equivalence sets. We use the extended version (\url{https://github.com/yao8839836/zsre}) and construct new locality sets following their methodology. Training data is sourced from the MEND project (\url{https://github.com/eric-mitchell/mend}).
    \item \textbf{WikiDatacounterfact}: This dataset focuses on triplets involving prominent Wikidata entities to mitigate issues with tail entities. Training data consists of randomly sampled Wikidata triplets, and the dataset itself serves as the test set.
    \item \textbf{WikiDatarecent}: A dataset of triplets recently added to Wikidata after July 2022, used to evaluate the insertion of new facts into models trained on older data.
\end{itemize}

\subsection{Metrics}

Knowledge editing affects predictions for inputs semantically or contextually related to the edited example. This sphere of influence is the editing scope. A successful edit modifies the model within the intended scope without affecting unrelated inputs:

\begin{equation}
f_{\theta_{e}}(x) = \left\{
\begin{array}{ll}
    y_{e} & \text{if } x \in I(x_{e}, y_{e}) \\
    f_{\theta}(x) & \text{if } x \in O(x_{e}, y_{e})
\end{array}
\right.
\end{equation}

where:

\begin{itemize}
    \item \(f_{\theta_{e}}(x)\): Edited model's prediction on input \(x\).
    \item \(y_{e}\): Target output for edited example \(x_{e}\).
    \item \(I(x_{e}, y_{e})\): Intended scope (inputs related to the edit).
    \item \(O(x_{e}, y_{e})\): Out-of-scope inputs (unrelated to the edit).
\end{itemize}

We evaluate edits using the following metrics:

\begin{itemize}
    \item \textbf{Edit Success (ES)}: Measures the model's accuracy on the edited fact and similar inputs (paraphrases) . For factual datasets, we use:
    \begin{equation}
    ES = \sum_{(x_k, y_k^*)}\mathbb{1}\{\text{argmax}_y f_{\theta'}(y | x_k) = y_k^*\}
    \end{equation}
    where \(x_k\) is the updated knowledge, \(y_k^*\) is the target output, and \(f_{\theta'}\) is the edited model.
    \item \textbf{Portability (PORT)}: Assesses the edit's impact on related knowledge, including alias/synonym substitution, compositionality/reasoning, and logical generalization.
    \item \textbf{Locality (LOC)}: Measures unintended changes to unrelated knowledge, considering both in-distribution and out-of-distribution locality:
    \begin{equation}
    \text{LOC} = \mathbb{E}_{x_k, y_k^* \sim O(x_k)}\mathbb{1}\{f_{\theta'}(y | x_k) = f_\theta(y | x_k)\}
    \end{equation}
    where \(O(x_k)\) represents unrelated knowledge, \(f_\theta\) is the original model, and \(f_{\theta'}\) is the edited model.
    \item \textbf{Fluency (FLUE)}: Assesses the edited model's generative capacity using the weighted average of bi-gram and tri-gram entropies. Lower values indicate higher repetitiveness.
\end{itemize}

These results confirm that the \textbf{Knowledge Neuronal Ensemble (KNE)} method provides excellent and stable editing performance across different datasets. It consistently delivers the best results in terms of editing accuracy, while maintaining high portability and locality metrics. In some cases, it even outperforms previous methods in specific datasets.

(Note: The results of the comparative knowledge editing methods were sourced from the repository: \href{https://github.com/zjunlp/EasyEdit?tab=readme-ov-file#editing-performance}{EasyEdit GitHub}.)

\section{The Figures of Experiment Results}

\begin{figure*}
	\centering
	\subfigure[]{\includegraphics[width=.4\linewidth]{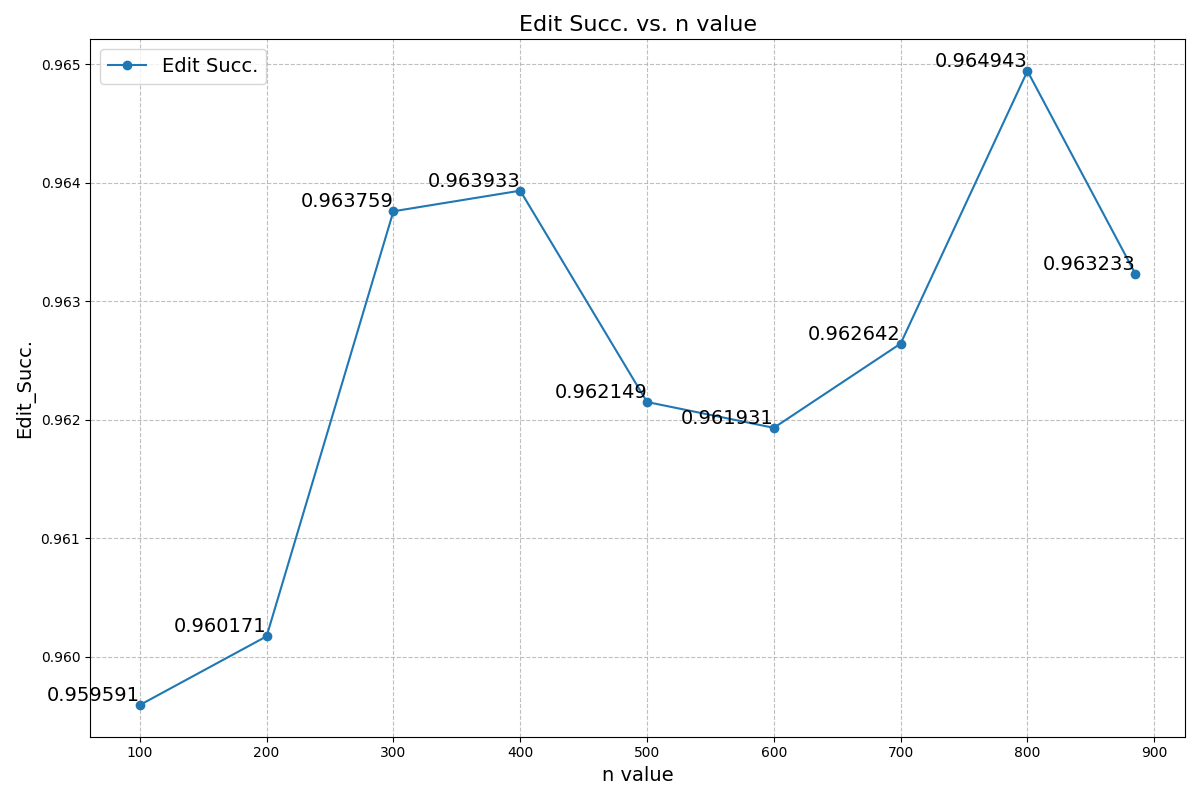}}
	\subfigure[]{\includegraphics[width=.4\linewidth]{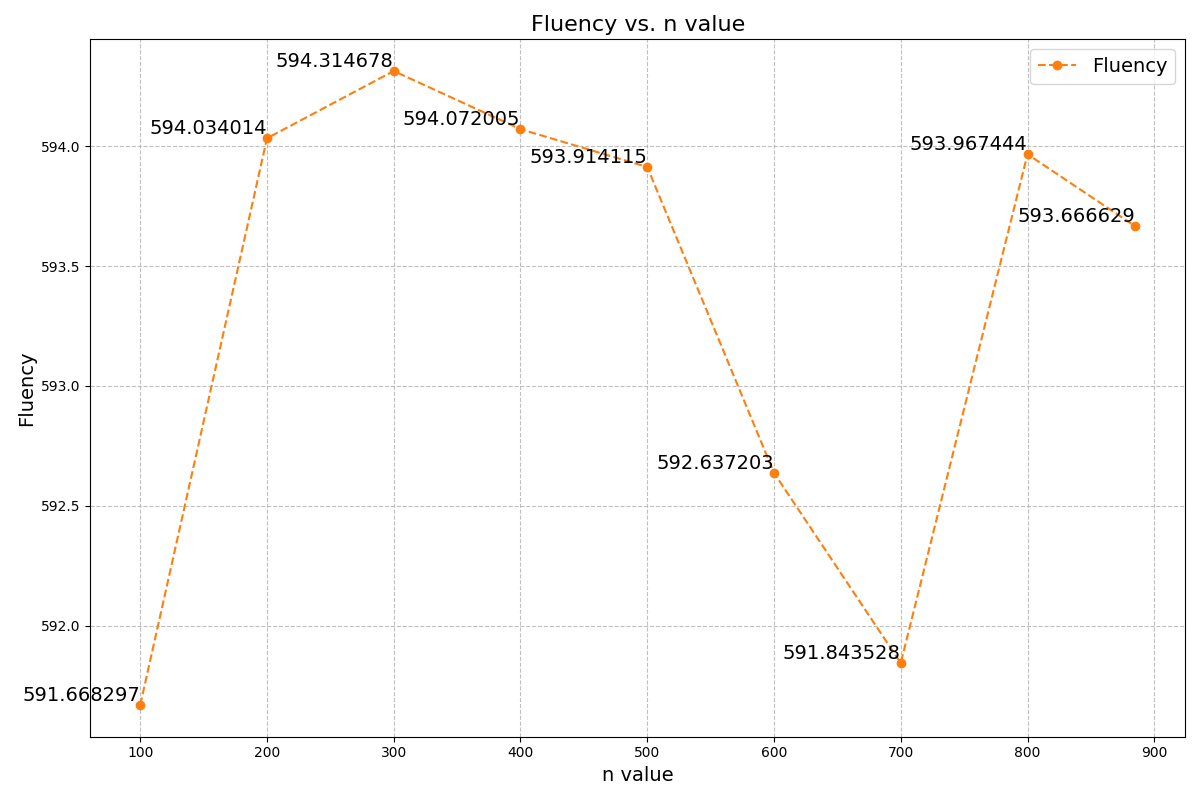}}
	\subfigure[]{\includegraphics[width=.4\linewidth]{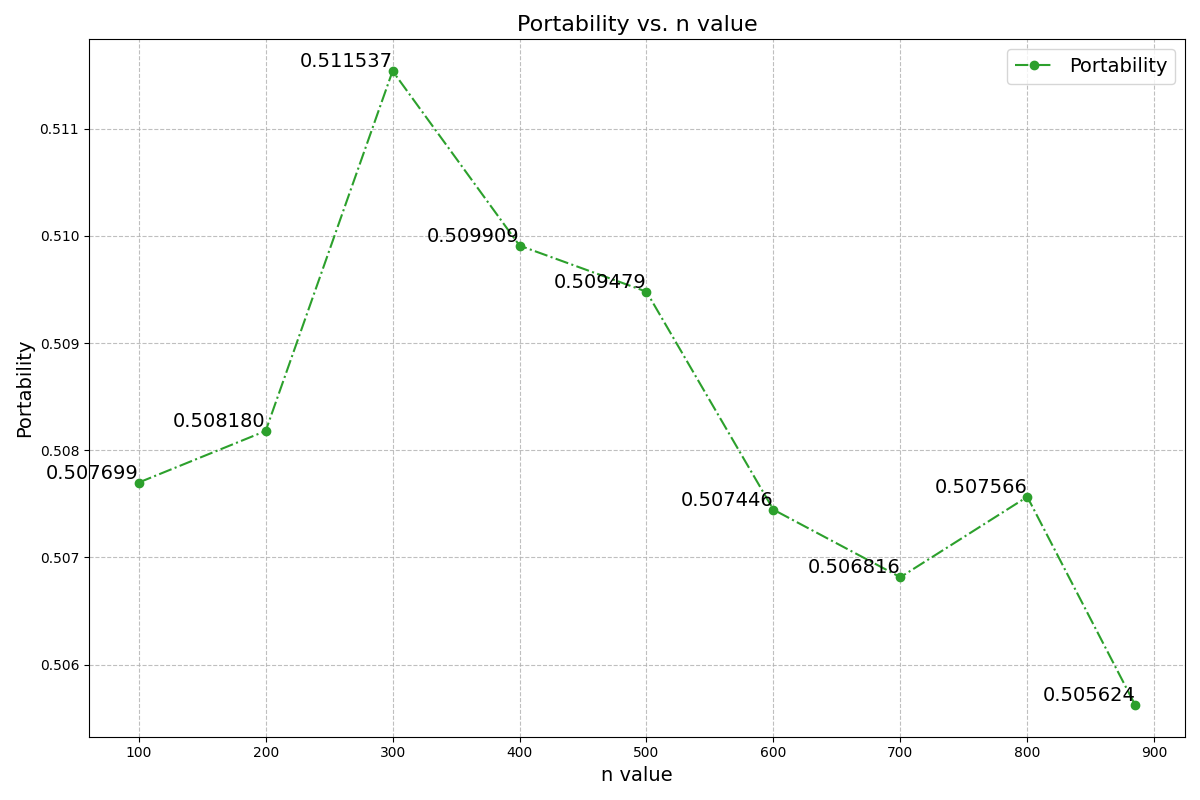}}
	\subfigure[]{\includegraphics[width=.4\linewidth]{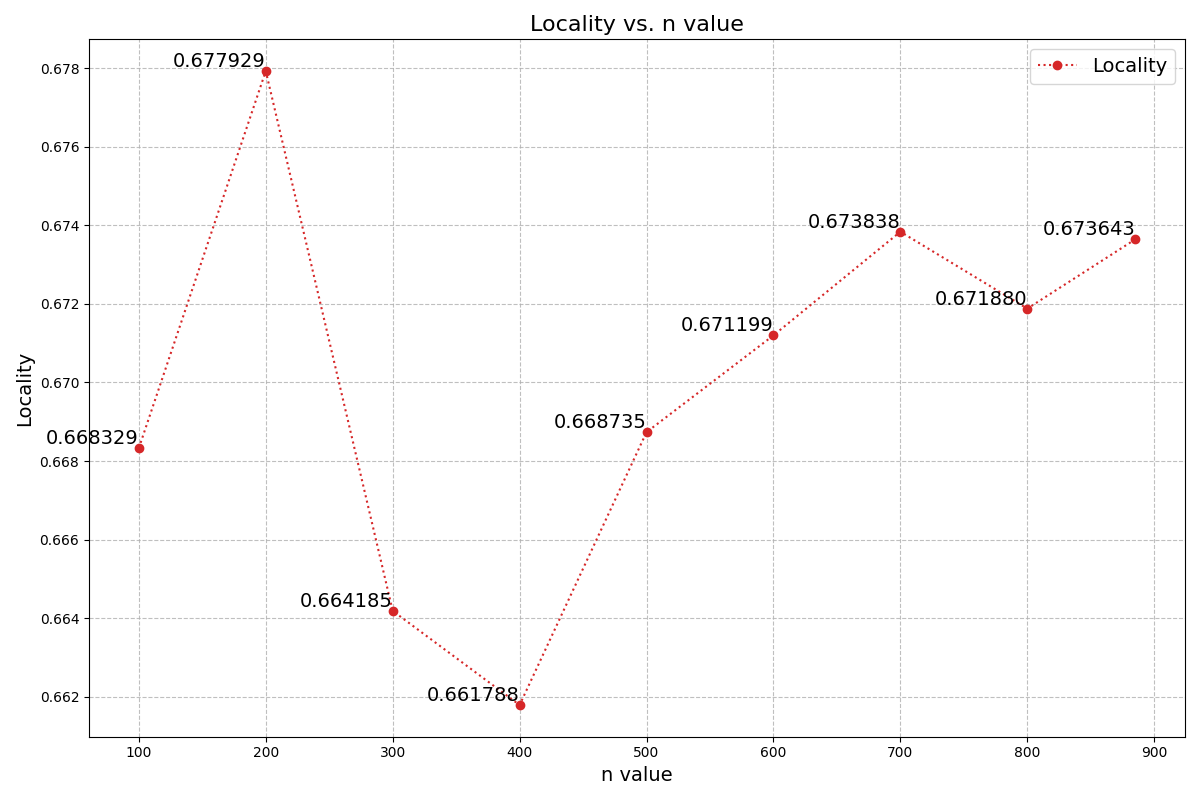}}
	\caption{Effects of Localized Knowledge versus Full Dataset Localization}
	\label{fig:full_dataset}
\end{figure*}

\begin{figure*}
	\centering
	\subfigure[]{\includegraphics[width=.4\linewidth]{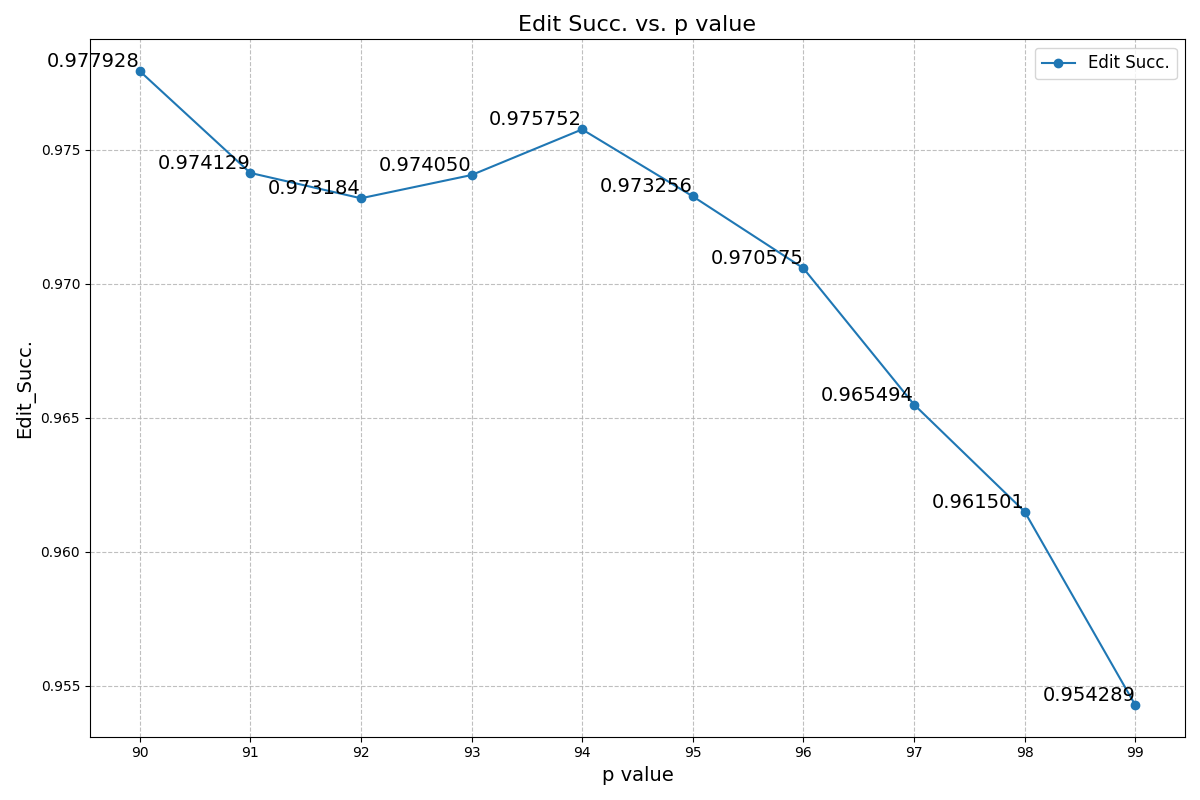}}
	\subfigure[]{\includegraphics[width=.4\linewidth]{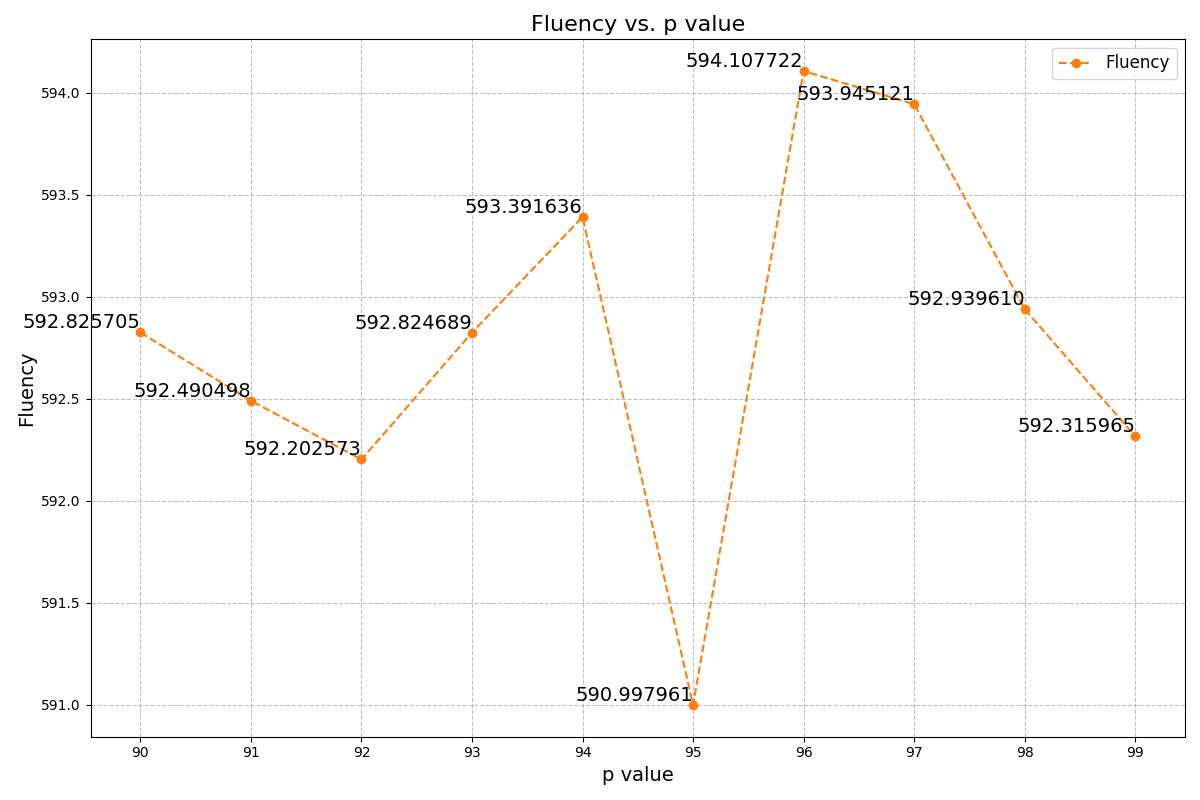}}
	\subfigure[]{\includegraphics[width=.4\linewidth]{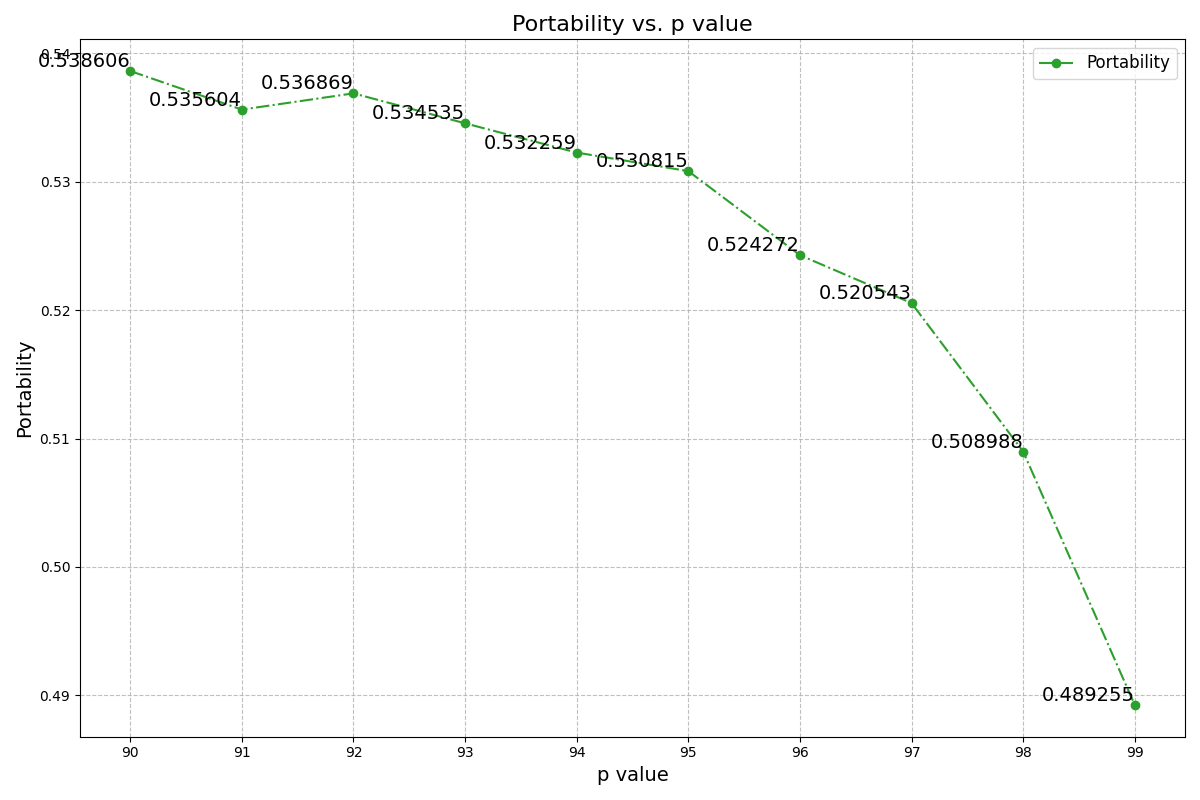}}
	\subfigure[]{\includegraphics[width=.4\linewidth]{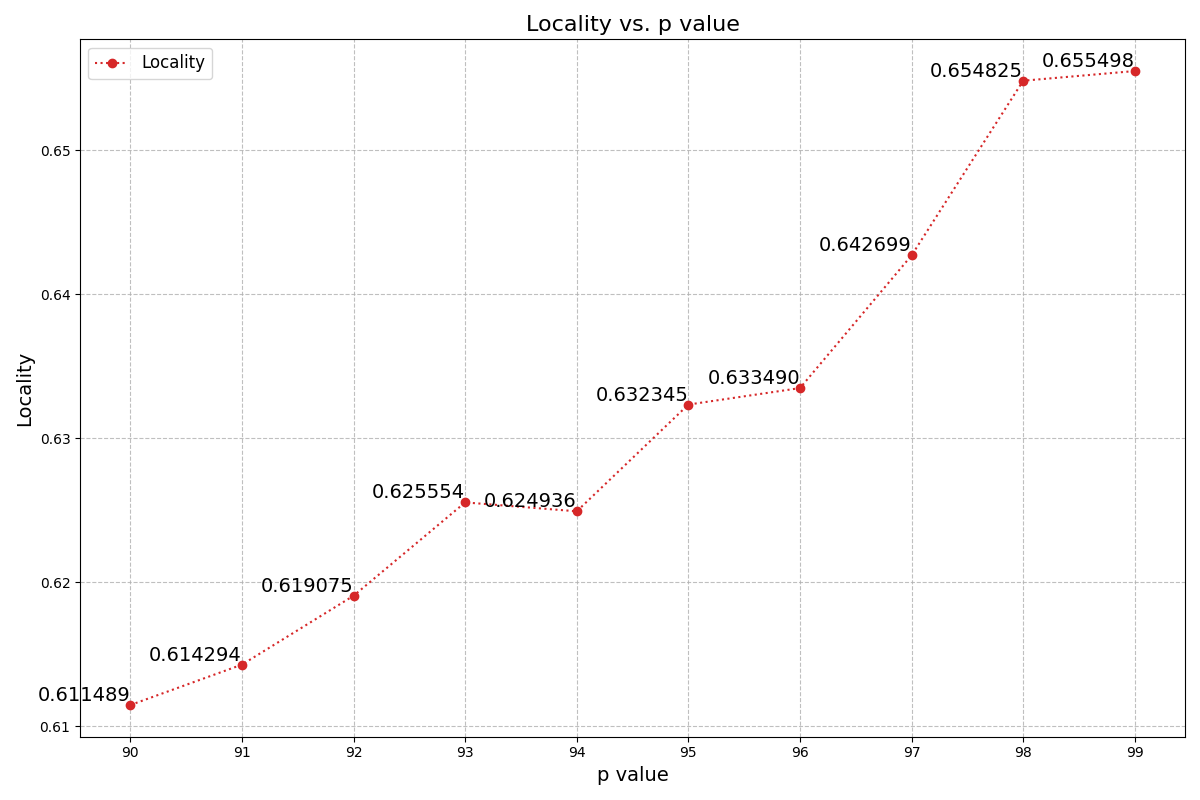}}
	\caption{Performance Metrics Across Varying Parameter Settings for Knowledge Editing}
	 \label{fig:parameter_setting}
\end{figure*}

\begin{figure*}
	\centering
	\subfigure[]{\includegraphics[width=.4\linewidth]{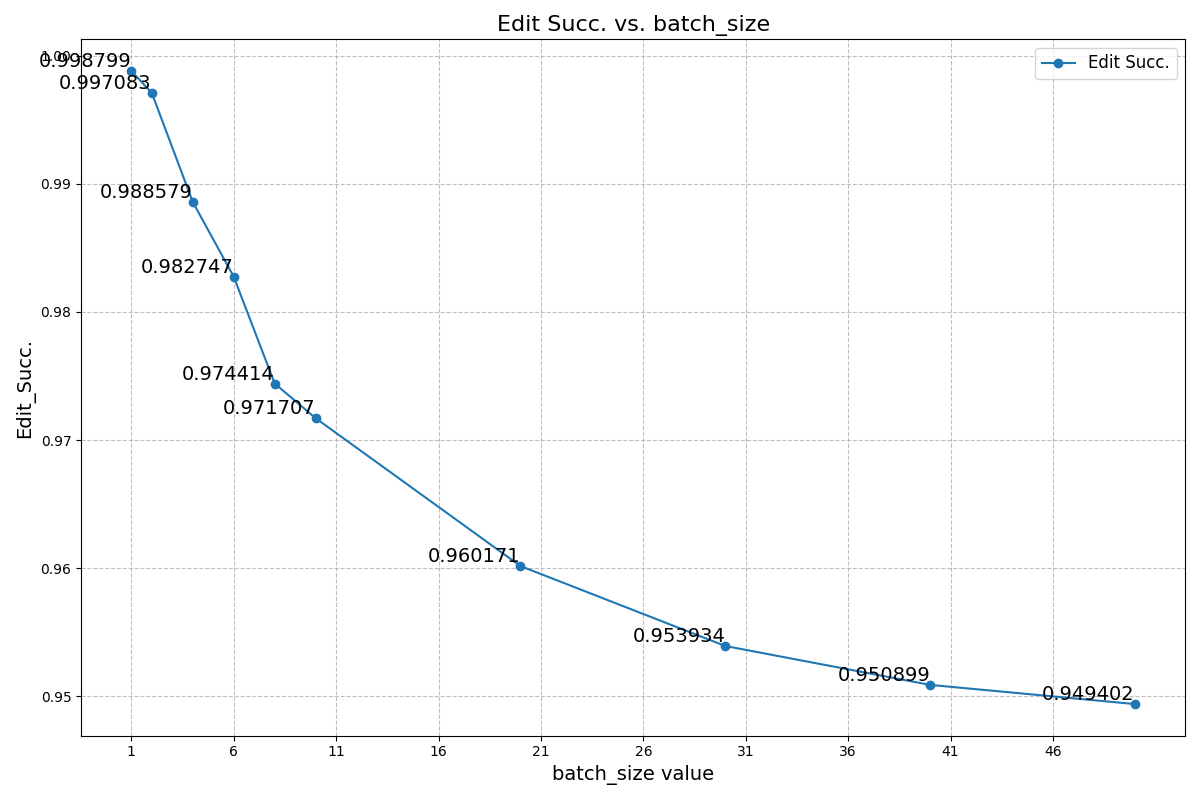}}
	\subfigure[]{\includegraphics[width=.4\linewidth]{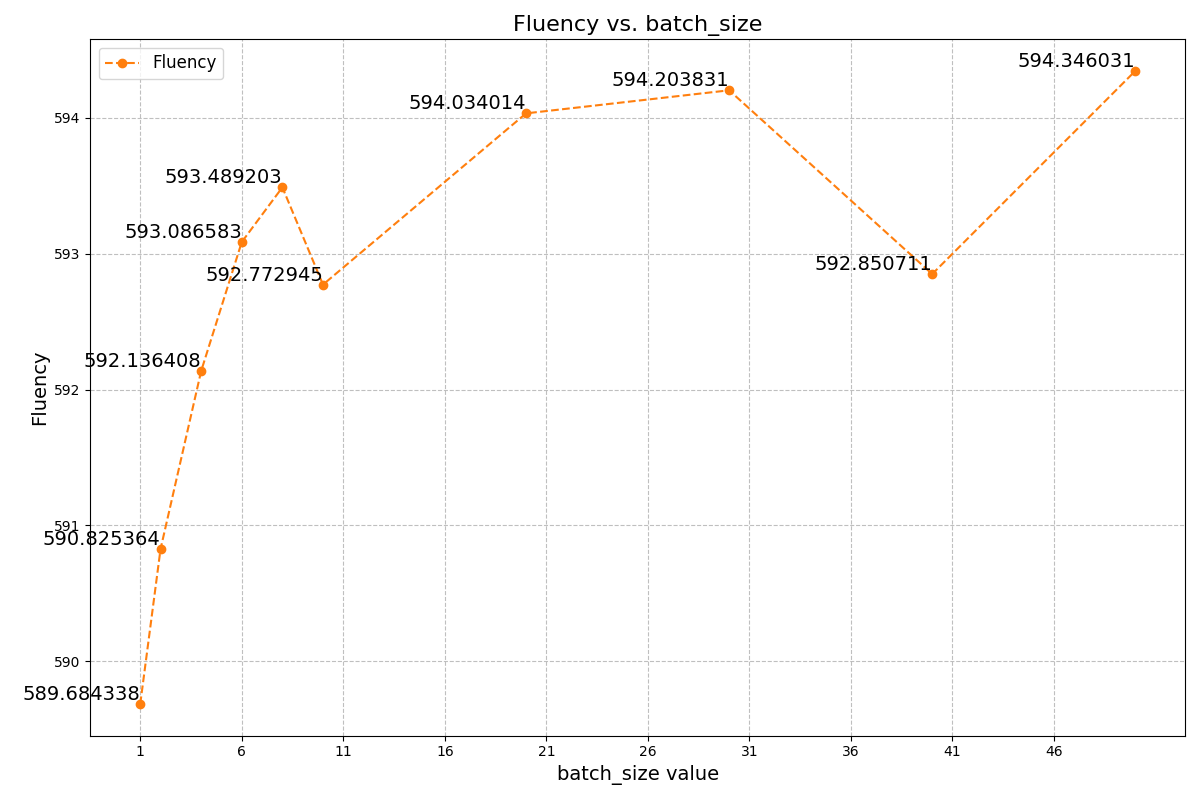}}
	\subfigure[]{\includegraphics[width=.4\linewidth]{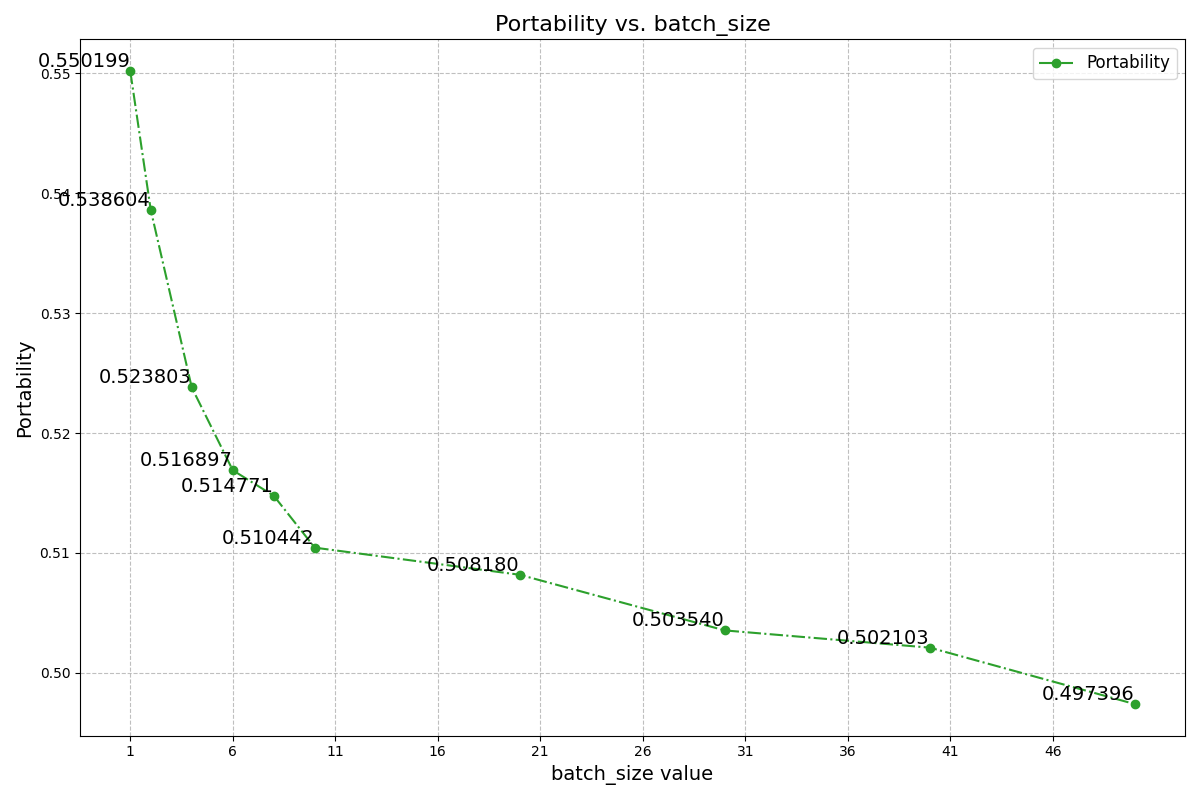}}
	\subfigure[]{\includegraphics[width=.4\linewidth]{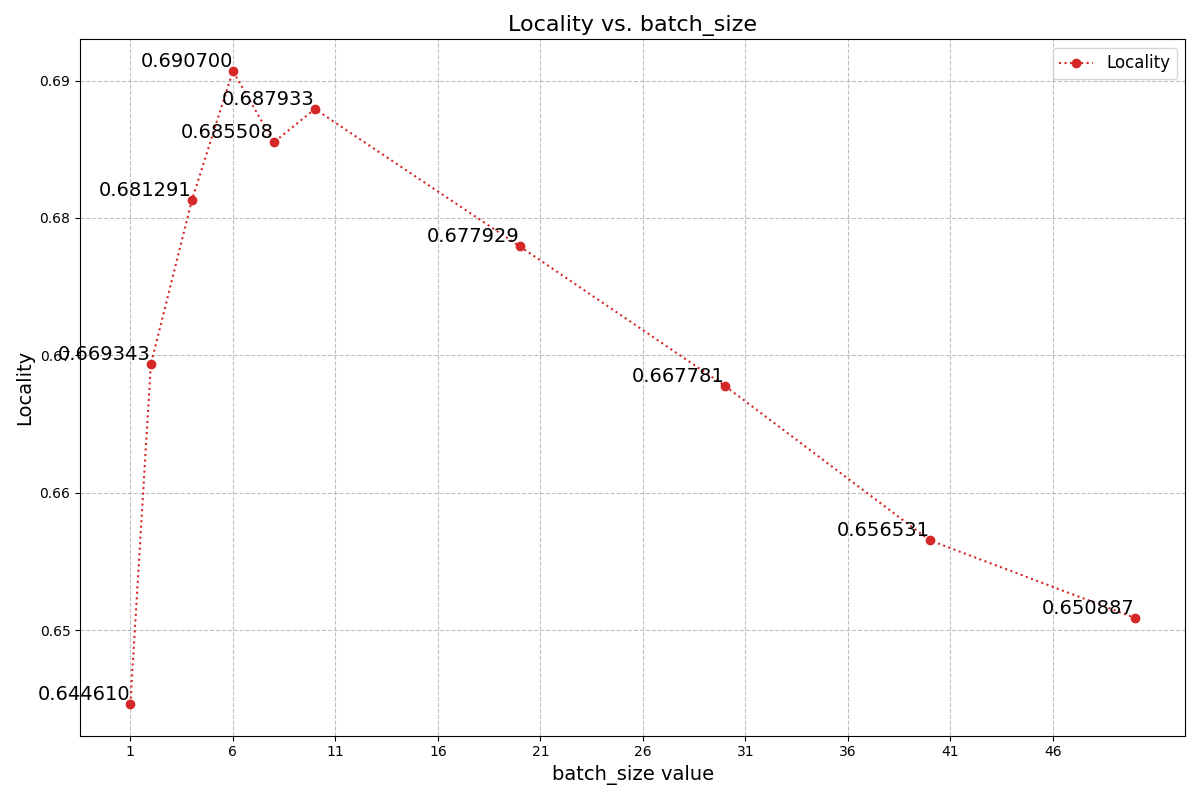}}
	\captionof{figure}{Performance Metrics Across Different Batch Sizes}
	\label{fig:batchsize}
\end{figure*}

\end{document}